# Advancing from Predictive Maintenance to Intelligent Maintenance with AI and IIoT

Haining Zheng
ExxonMobil Research and
Engineering Company
Annandale, NJ 08801
haining.zheng@exxonmobil.com

Antonio R. Paiva
ExxonMobil Research and
Engineering Company
Annandale, NJ 08801
antonio.paiva@exxonmobil.com

Chris S. Gurciullo
ExxonMobil Research and
Engineering Company
Annandale, NJ 08801
chris.s.gurciullo@exxonmobil.com

## ABSTRACT

As Artificial Intelligent (AI) technology advances and increasingly large amounts of data become readily available via various Industrial Internet of Things (IIoT) projects, we evaluate the state of the art of predictive maintenance approaches and propose our innovative framework to improve the current practice. The paper first reviews the evolution of reliability modelling technology in the past 90 years and discusses major technologies developed in industry and academia. We then introduce the next generation maintenance framework - Intelligent Maintenance, and discuss its key components. This AI and IIoT based Intelligent Maintenance framework is composed of (1) latest machine learning algorithms including probabilistic reliability modelling with deep learning, (2) real-time data collection, transfer, and storage through wireless smart sensors, (3) Big Data technologies, (4) continuously integration and deployment of machine learning models, (5) mobile device and AR/VR applications for fast and better decision-making in the field. Particularly, we proposed a novel probabilistic deep learning reliability modelling approach and demonstrate it in the Turbofan Engine Degradation Dataset.

## 1 Introduction

Equipment reliability has been a major issue for manufacturers of many industries. Based on Aberdeen's independent research of unplanned downtime for industrial plants costs $10k to $250K/hour which adds up to $50 billion annually. Equipment failure is the cause of 42% of this unplanned downtime [1]. While reliability technology has been studied intensively in the past 90 years [2-5], a 2017 survey of 100 manufacturers in the US and Europe by Vanson Bourne Global Study shows that 70% of companies lack complete awareness of when equipment is due for maintenance or upgrade [6].

Machine learning [7-9] and the Internet of Things (IoT) [10-12] have made significant progress recently and proven successful across different industries, including a number of traditional applications in the energy industry [13-15] from upstream production prediction, midstream transportation optimization, to downstream product manufacturing. Equipped with our years of experience in the reliability technology domain and knowledge of latest AI and IIoT development, we revisit this important yet unsolved problem.

## 2 Evolution of Reliability Technology

Reliability technology has gone through four generations as illustrated in Figure 1.

### 2.1 Reactive Maintenance

The first generation of maintenance strategy follows a reactive approach: only fix it when the equipment is broken. This caused tremendous unplanned capacity loss because it is subject to availability of repair personnel and more often than not, the asset is severely damaged to the point that it needs to be replaced entirely which is costly and subject to replacement parts availability. Currently it is only applied to inexpensive and easily replaceable small assets for which spare parts can be easily kept in stock.

### 2.2 Preventive Maintenance

After World War II, a 2[nd] generation of maintenance strategies were developed, giving rise to preventive maintenance, meaning that replacement of equipment is scheduled according to a fixed time interval, regardless of the condition. Of course, this approach creates a major dilemma for business decision makers: either they apply a large safety factor to serve and replace equipment frequently which increases maintenance costs, or they face situations in which the asset breaks before its expected lifespan causing unplanned capacity loss similar to reactive maintenance situation.

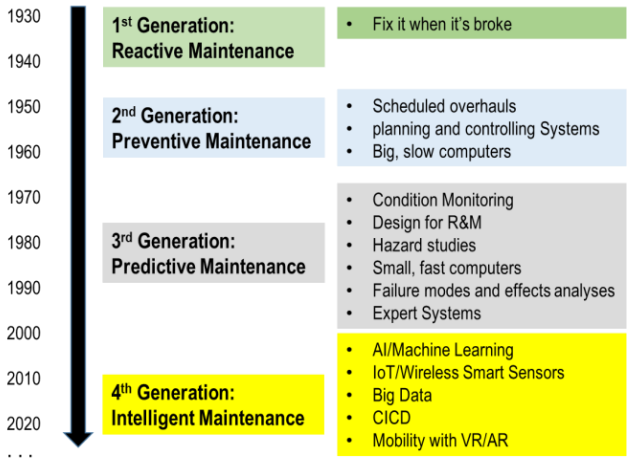

**Figure 1: Evolution of Reliability Technology (adapted from Moubray [2]).**

### 2.3 Predictive Maintenance

As light-weight and fast computers became widely available in the 1980s, predictive maintenance becomes practical. The goal is to preemptively predict equipment failure through data from conditional monitoring and computer models. A number of reliability technologies were developed in this period. In the equipment/asset development phase, Accelerated Testing, Design for Reliability and Maintenance, and Design Failure Mode Analysis (DFMEA) are important design supporting tools. During the project and operational phases, Reliability Centered Maintenance (RCM), Reliability-Based Inspection (ReBI), Optimum Replacement Time (ORT), and Reliability, Availability, and Maintainability (RAM) analysis are commonly used approaches. Moreover, Fault Tree Analysis (FTA) and event tree analysis (ETA) were developed for failure diagnosis (root cause analysis) to identify the main causes of failure of an asset after a failure has occurred. Rule-based expert systems were also developed in this period.

An expansion of predictive maintenance is prescriptive maintenance which emphases on quantify/predict the effect of maintenance decisions before they are made. Its goal is to recommend what mitigation or maintenance actions and by when they need to be done on an asset. Predictive Maintenance is the most prevalent strategy presently. Currently 50% of the manufacturers have established continuous improvement teams for condition based and RCM activities [1].

### 2.4 Intelligent Maintenance

While Industry 4.0 is revolutionizing every aspect of industrial process and turning what was unimaginable into reality, like real time plant-wide optimization and scheduling, manufacturing system is becoming increasingly more complex and brings in new challenges to maintenance strategy.

(1) How to implement the AI/Machine learning algorithms developed in other fields to manufacturing time series data which has complex nonlinear temporal and spatial dynamics?

(2) How to collect data from remote sources that are not connected to the corporate network via wired connections?

(3) How to effective process and store high frequency data without overloading the whole network or data storage?

(4) How to keep models deployed up to date without causing disruptions to production?

(5) How to allow fast and better decision-making in the field without access to laptop or workstation?

In the next section, we introduce a framework needed to tackle these challenges and advance towards Intelligent Maintenance and discuss the barriers and opportunities for practical implementation.

## 3   Intelligent Maintenance Framework

This Intelligent Maintenance framework is composed of five elements as shown in the yellow colored part of Figure 1 and which we will discuss in detail in this section.

### 3.1 AI/Machine Learning Applied to Reliability

3.3.1 Supervised Learning

Regression and classification are two most common Supervised Learning approaches. For equipment failure prediction, a regression formulation can be employed to predict when an in-service machine will fail in the future, so that maintenance can be planned in advance. Estimates of Remaining Useful Life (RUL) and Time to Failure (TTF) are the most common regression targets. Machine learning models employed toward that end include Boosted Decision Trees Regression, Random Forest Regression, Poisson Regression, and Neural Network-based Regression.

On the other hand, the same equipment failure prediction problem can be framed as classification: either as a Binary Classification to predict if an asset will fail within certain time window (e.g., 30 days), or as a Multi-class Classification to predict if an asset will fail in one of different time windows: e.g., fails in window $[0, w_0]$ days, fails in the window $[w_0+1, w_1]$ days, … fails in the window $[w_n+1, w_{n+1}]$ days and not fail within $w_{n+1}$ days. Logistic Regression, Support Vector Machine (SVM), Decision Tree, Random Forrest, eXtreme Gradient Boosting (XGBoost) and Neural Network are common algorithms. In addition, classification models can help identify failure types.

3.3.2 Unsupervised Learning

Anomaly detection is the most common unsupervised learning framework for maintenance analytics. It's used to detect anomalies in equipment or system performance or functionality. K-means, Isolation Forrest, Local Outlier Factor (LOF) are mostly commonly used models.

3.3.3 Probabilistic approach

As Data Science practitioners, we are facing significant challenges arising from actual manufacturing systems, such as:

(1) Low accuracy of the maintenance records (labels), and large number of undocumented and "unrelated" shutdown and maintenance events add noise to the data.

(2) Run to failure is rare due to the high cost of unplanned capability loss. Thus the decision to conduct maintenance is often complex and involves substantial experience, engineering, and logistical judgement. It could be performed long after the event occurred due to lack of parts or before any event as a preventive measure (planned maintenance), or sometimes multiple parts are replaced together based on opportunity and engineering judgement. This further increases difficulty to collect data and interpolate data correctly.

(3) High-degree of nonlinear temporal and multidimensional correlations between different types of upstream and downstream sensor data.

(4) The fundamental physics behind the manufacturing system is highly nonlinear and non-explicit.

(5) Process and sensor variables are often non-Gaussian distributed, which prevents simple statistical analysis and methods.

(6) The normal operating condition, which defines the baseline for anomaly detection algorithms, is constantly changing and difficult to define even with a domain expert's help.

Thus, Probabilistic approaches, such as a recently developed Bayesian recurrent neural network (BRNN) architecture [16-18], can help address a few key aspects of these challenges and serve as an example to demonstrate how latest development of machine learning can help advance predictive maintenance in section 4.

**3.2 IIoT and Smart Sensors**

The fast development of Industrial IoT helps achieve real-time data acquisition and connect isolated data source to corporate network with wireless sensors. Edge computing technology enables model building and its deployment at distributed IoT edge devices. Nevertheless, even the best-in-class companies, only 30% of them have IIoT platforms to collect device data, build smart apps and enable industrial scale analytics for application performance management while the laggards were at 10% in 2017 [1].

**3.3 Big Data Analytics**

Big Data are usually branded using the famous 3 Vs (Volume, Velocity and Variety) – large volume of data streaming at high velocity with different varieties of datatypes from relational databases to unstructured and semi-structured data. Hadoop Data Lake serves as the primary repository for incoming streams of raw data and data is stored in the Hadoop Distributed File System (HDFS) after being processed by extract, transform and load (ETL) integration jobs. Then the data can be used for advanced analytics by running through a processing engine like Spark, which enables users to run large-scale data analytics applications across clustered systems in parallel.

While large organizations mostly deployed Big Data system on premises, particularly in early days, public cloud platform vendors, such as Microsoft Azure, Amazon Web Services (AWS) and Google Cloud Platform (GCP) have each obtained significant market share. Hybrid Cloud is a recent trend with mixed computing, storage, and services environment made up of on-premises infrastructure, private cloud services, and public cloud services with benefits of security, control, agility and cost.

**3.4 Continuous Integration and Continuous Deployment**

Continuous Integration and Continuous Deployment (CICD) helps to keep models deployed up to date without causing disruptions to production. While Continuous Integration (CI) establishes a consistent and automated way to build, package, and test applications, Continuous delivery (CD) automates the delivery of applications to selected environments (development, testing or production). CI/CD automation performs any necessary service calls to web servers, databases, and other services and keeps the deployed Machine Leaning models up to date without causing disruptions to production.

**3.5 Mobility and VR/AR**

Mobile devices provide engineers access to job orders, equipment statics, machine schematics and part inventory in real time and instantaneous visualizations in the field, such that they can make fast and enhanced decisions without traveling back and forth between offices and field.

One step further, Virtual Reality (VR) and Augmented Reality (AR) technologies can simulate key processes and performing virtual tests of production lines and equipment and helps pinpoint mistakes which could lead to potential disruptions and eliminate them before they stall operations.

## 4 Case Study

We now demonstrate how to leverage recent advances in machine learning discussed in section 3.1 and increasing levels of sensor data from IIoT networks discussed in section 3.2 to help advance from predictive maintenance to intelligent maintenance and tackle several aspects of the previously mentioned challenges.

### 4.1 Approach

In the following, we use the recently proposed Bayesian recurrent neural network (BRNN) framework with variational dropout [15-17]. BRNNs model the joint distribution and nonlinear complex dynamics between all variables (i.e., machine settings and sensor measurements). Through variational dropout, BRNNs yield estimates of the prediction uncertainty, which capture both model uncertainty and the inherent noise in the data. These mean that BRNN modeling tackles four of the aforementioned challenges: (1) nonlinear spatio-temporal correlations (i.e., correlations between variables and time lags), (2) non-Gaussianity in system variables, (3) characterization of uncertainty in the predictions, and (4) data-driven modeling, without the need for explicit models of the system. Furthermore, the dropout technique used in the training of these models inherently regularizes and improves the robustness of the predictions.

BRNNs were implemented using the dropout technique at both training and testing. The goal is use dropout as a variational approximation for efficient inference. For predictions, this involves drawing samples of the model and evaluating each model sample. The model samples are obtained by applying dropout which drops randomly selected inputs, outputs, and hidden states. This results in multiple random realizations of the RNN model, each obtained by implicitly removing a portion of the inputs, outputs, or hidden states. One can then collect statistics over the predictive distribution, which characterize uncertainty in the model predictions. This approach can be implemented in Keras with TensorFlow rather easily.

Specifically, the BRNN models used comprised 2 layers of LSTM nodes, each with 100 and 50 nodes respectively, and followed by a single-output dense layer as illustrated in Figure 2. Dropout was applied in 3 places: (1) in the input of second LSTM layer with a drop rate of 10%, (2) in the states between time lags of both LSTM layers with a drop rate of 10%, and (3) in the inputs of the final dense layer with a drop rate of 20%. The network training used time-sequences of 50 samples and the Adam optimizer. Stopping was determined by early-stopping on a 10% validation set.

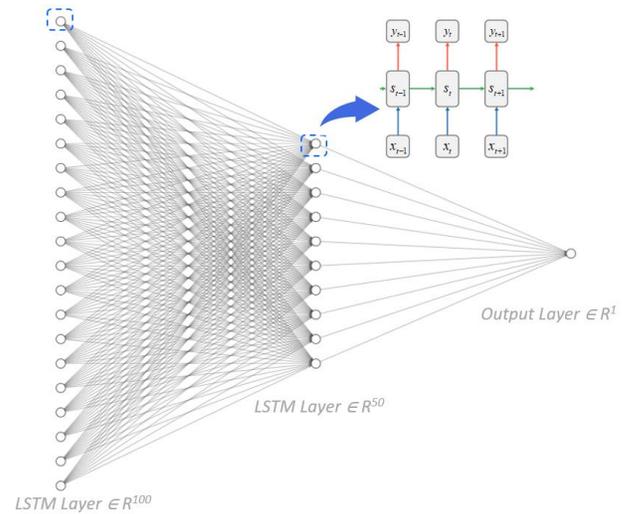

**Figure 2. BRNN model neural net architecture schematics.**

### 4.2 Turbofan Engine Degradation Dataset [19]

This methodology is demonstrated using a turbofan engine degradation dataset. Since the Turbofan Engine is highly expensive to fix, predicting its time to failure (TTF) can help prevent turbofan failures, and minimize downtime. Failure probabilities will inform technicians to monitor turbofan engines that are likely to fail soon, and schedule maintenance regimes.

IIoT sensors monitor the status of critical operating components by recording vibration, temperature and pressure. Data can be gathered from multiple turbofan engines in various regions and transmitted to cloud for batch processing and further predictive analytics.

The dataset used here was obtained by simulation using C-MAPSS and was used as benchmark in the challenge competition held at the 1st international conference on Prognostics and Health Management (PHM08). Four different datasets were simulated under a number of combinations of operating conditions and failure modes. The recordings include a total of 24 variables recordings: 3 settings variables (i.e., system inputs) and 21 sensor variables to characterize the "system health". All test sequences were stopped before the actual failure event, but the time-to-failure was recorded.

The problem was formulated as a binary classification task corresponding to whether a failure is likely to occur within the next 30 days, as discussed in section 3.1.1.

### 4.3 Results

The performance of this method is demonstrated and contrasted to a standard RNN, which are widely applied to process time series predictions in the industry. Figure presents a test prediction example an engine that was going to fail. Although the simulation was halted before failure, the yellow shaded area indicates the warning time window that a failure would occur in 30 or less days. Note that the BRNN yield a predictive distribution, shown in the figure using shaded bands corresponding to the 10 to 90 percentile range and 25 to 75 percentile range. The center curve is the median predicted probability of failure.

By aligning with respect to the warning window, we similarly show the test predictive distribution results for engines known to be approaching failure in Figure . The results show that BRNN robustly indicates with high probability when the engines are approaching the failure.

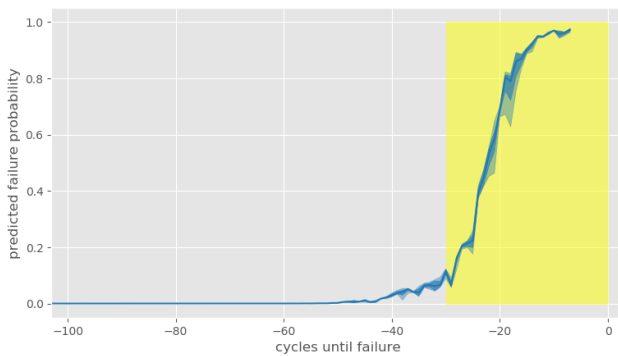

**Figure 3: BRNN distribution of probability of failure for an engine that would fail. The shaded region denotes the 30 day to failure time window.**

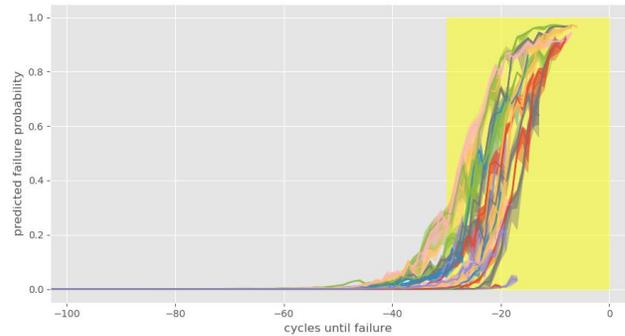

**Figure 4: BRNN distribution of probability of failure for all test engines known to be approaching failure.**

For comparison, the results using a standard RNN model and with similar architecture are shown in Figure . There is increased spread between the curves and the probability of failure increases before the 30 day time window for a number of engines. Thus, maintenance decisions based on these predictions would likely lead to a larger number/more frequent maintenance events, thereby reducing slightly the uplift of the predictive maintenance system.

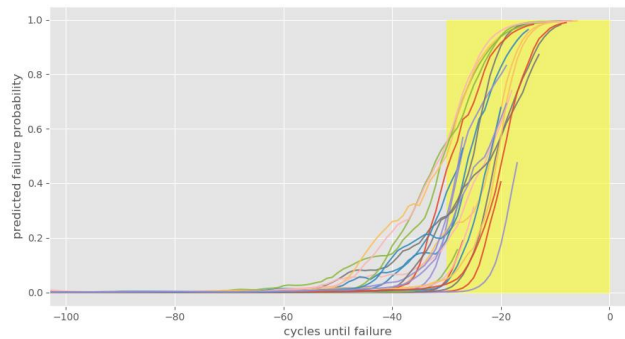

**Figure 5: Probability of failure for an engine predicted by standard RNN.**

### 4.4 Discussion

The case study shows how advances in BRNNs can help advance current predictive maintenance practices. There are still a number of open challenges however. The example relied on labels of when the failure occurred. How to learn and infer the need for maintenance with perhaps only a few examples or without even letting reach that point remains a research topic. One approach could involve an increased role of survival analysis models ([20] but they still need enough examples to learn the underlying model feature representation. Another recent development that may help alleviate this issue are self-supervised learning techniques [21] for learning the model feature representation without labels.

Yet another consideration toward Intelligent Maintenance is that prediction of the need for maintenance is only the beginning. As previously mentioned, in practice many other considerations need to be taken into account when managing an inventory of machines, such as logistic considerations and parts availability. Thus, the result of the predictive maintenance model should be input for schedule management optimization such as to minimize overall risk of failure. It is in that respect that uncertainty in the predictions of the model play a major role because they provide an optimizer with the ability to estimate how eminent is the failure.

Finally, the goal of these systems is to inform operators, which often have many years of experience and engineering judgement. Hence, the ability to understand the why for the guidance from such systems is crucial. In that respect, neural network-based methods are notoriously opaque, but this is a very active area of research [22].

## 5 Conclusions

In this paper, we reviewed past maintenance strategy and discussed the evolution of reliability technology. Benefiting from the development of Artificial Intelligent and Industrial IoT technology, we introduced the next generation maintenance framework, Intelligent Maintenance, and discussed its key components. It's a AI and IIoT based maintenance framework that combines the real-time data collection, transfer, and storage through wireless sensors and Big Data technologies, continuously train and deploy the machine learning models, and implementation at mobile device, as well as AR/VR, for fast and better decision making in the field. Finally, a case study was presented as example of methods that will enable the above framework. With 72% of organizations considering zero unplanned downtime as the No. 1 priority or a high priority[6], advancing from Predictive Maintenance to Intelligent Maintenance with AI and IIoT is a solid step for the ultimate goal of autonomous running manufacturing lines 24/7 with zero downtime in future enterprise.

## ACKNOWLEDGMENTS

The authors would like acknowledge helpful discussions and support from Sasha Mitarai, Prasenjeet Ghosh, Peng Xu, Timothy Westhoven, Liezhong Gong, Jeff Ludwig and Anantha Sundaram.

## REFERENCES

[1] Aberdeen report (2016), *Maintaining Virtual System Uptime in Today's Transforming IT Infrastructure*
[2] J. Moubray (2001). *Reliability centered maintenance*, Industrial Press Inc.
[3] A. K. S. Jardine, A. H. C. Tsang (2013). *Maintenance, replacement, and reliability: theory and applications*, CRC press.
[4] W. Q. Meeker, L. A. Escobar (1998), *Statistical Methods for Reliability Data*, Wiley-Interscience
[5] R. K. Mobley (2002). *An Introduction to Predictive Maintenance*, Butterworth-Heinemann.
[6] Vanson Bourne global study (2017), *After the Fall: Cost, Causes and Consequences of Unplanned Downtime.*
[7] C. M. Bishop (2016), *Pattern Recognition and Machine Learning*, Springer-Verlag New York
[8] T. Hastie (2009), *The Elements of Statistical Learning: Data Mining, Inference, and Prediction*, Springer-Verlag, New York
[9] I. Goodfellow, Y. Bengio, A. Courville (2016). *Deep Learning*, MIT Press
[10] A. Colakovic, M. Hadzialic (2018).*Internet of Things (IoT): A review of enabling technologies, challenges, and open research issues*. Computer Networks, 144, 17-39.
[11] D. Singh, G. Tripathi, A. J. Jara (2014). *A survey of internet-of-things: future vision, architecture, challenges and services.* Proc. IEEE World Forum on Internet of Things, 287-292.
[12] J. Gubbi, R. Buyya, S. Marusic, M. Palaniswami (2013). *Internet of Things (IoT): a vision, architectural elements, and future directions.* Future Generation Computer Systems. February 1645-1660.
[13] W. Z. Khan, M. Y. Aalsalem, M. K. Khan, M. S. Hossain, M. Atiquzzaman (2017). *A Reliable Internet of Things based Architecture for Oil and Gas Industry.* International Conference on Advanced Communications Technology, 19-22.
[14] M. Y. Aalsalem, W. Z. Khan, W. Gharibi, M. K. Khan, Q. Arshad (2018). *Wireless Sensor Networks in oil and gas industry: Recent advances, taxonomy, requirements, and open challenges*. Journal of Network and Computer Applications, 113, 87-97.
[15] N. Awalgaonkar, H. Zheng, C. Gurciullo, "DEEVA: A Deep Learning and IoT Based Computer Vision System to Address Safety and Security of Production Sites in Energy Industry", The Thirty-Fourth AAAI Conference on Artificial Intelligence (AAAI-20) Artificial Intelligence of Things (AIoT) Workshop, New York, NY, Feb. 7-11, 2020 http://arxiv.org/abs/2003.01196
[16] Y. Gal, Z. Ghahramani (2016a), *A theoretically grounded application of dropout in recurrent neural networks,* in Advances in Neural Information Processing Systems, pp. 1019-1027.
[17] Y. Gal, Z. Ghahramani (2016b), *Dropout as a Bayesian approximation: Representing model uncertainty in deep learning,* in Proceedings of the International Conference on Machine Learning, pp.1050-1059.
[18] W. Sun, A. R. Paiva, P. Xu, A. Sundaram, R. D. Braatz (2020), *Fault Detection and Identification using Bayesian Recurrent Neural Networks, Computers & Chemical Engineering, Volume 141, 106991, https://doi.org/10.1016/j.compchemeng.2020.106991*
[19] Turbofan Engine Degradation Simulation Data Set, NASA Ames Prognostics Data Repository (https://ti.arc.nasa.gov/tech/dash/groups/pcoe/prognostic-data-repository/#turbofan), NASA Ames Research Center, Moffett Field, CA
[20] D. G. Kleinbaum, M. Klein, (2010), *Survival Analysis*, Springer.
[21] C. Doersch, A. Gupta, A. A. Efros (2015). *Unsupervised visual representation learning by context prediction*. In Proceedings of the IEEE International Conference on Computer Vision (pp. 1422-1430).
[22] F. Fan, J. Xiong, G. Wang (2020). *On Interpretability of Artificial Neural Networks. arXiv preprint arXiv:2001.02522*.